\documentclass[10pt,twocolumn,letterpaper]{article}

\usepackage{cvpr}
\usepackage{times}
\usepackage{epsfig}
\usepackage{graphicx}
\usepackage{amsmath}
\usepackage{amssymb}

\usepackage{threeparttable}
\usepackage{algorithm}
\usepackage{algorithmic}
\usepackage{multirow}
\usepackage{breqn}
\usepackage{subfigure}
\usepackage{diagbox}
\graphicspath{{../figures/}}

\def\K{{\bf K}}

\def\kk{{\bf k}}

\def\I{{\bf I}}
\def\R{{\bf R}}
\def\X{{\bf X}}
\def\Y{{\bf Y}}

\def\x{{\bf x}}

\def\u{{\bf u}}
\def\V{{\bf V}}
\def\vv{{\bf v}}
\def\W{{\bf W}}

\def\0{{\bf 0}}
\def\1{{\bf 1}}

\def\DM{{\mathcal D}}

\def\KM{{\mathcal K}}
\def\LM{{\mathcal L}}
\def\TM{{\mathcal T}}
\def\UM{{\mathcal U}}
\def\XM{{\mathcal X}}

\def\SM{{\mathcal S}}

\def\RB{{\mathbb R}}

\def\alp{\mbox{\boldmath$\alpha$\unboldmath}}
\def\bet{\mbox{\boldmath$\beta$\unboldmath}}

\def\eg{\emph{e.g. }}
\def\ie{\emph{i.e. }}

\usepackage[breaklinks=true,bookmarks=false]{hyperref}

\cvprfinalcopy 

\def\cvprPaperID{1224} 

\begin{document}

\title{Zero-Shot Recognition using Dual Visual-Semantic Mapping Paths}

\author{Yanan Li \qquad Donghui Wang\thanks{Corresponding author} \qquad Huanhang Hu \qquad Yuetan Lin \qquad Yueting Zhuang\\
Institute of Artificial Intelligence, Zhejiang University\\
{\tt\small \{ynli, dhwang, huhh, linyuetan, yzhuang\}@zju.edu.cn}
}

\maketitle

\begin{abstract}
  Zero-shot recognition aims to accurately recognize objects of unseen classes by using a shared visual-semantic mapping between the image feature space and the semantic embedding space. This  mapping is learned on training data of seen classes and is expected to have transfer ability to unseen classes. In this paper, we tackle this problem by exploiting the intrinsic relationship between the semantic space manifold and the transfer ability of visual-semantic mapping. We formalize their connection and cast zero-shot recognition as a joint optimization problem. Motivated by this, we propose a novel framework for zero-shot recognition, which contains dual visual-semantic mapping paths. Our analysis shows this framework can not only apply prior semantic knowledge to infer underlying semantic manifold in the image feature space, but also generate optimized semantic embedding space, which can enhance the transfer ability of the visual-semantic mapping to unseen classes. The proposed method is evaluated for zero-shot recognition on four benchmark datasets, achieving outstanding results.
\end{abstract}

\section{Introduction}
\label{sec:introduction}
Visual object recognition typically requires a large collection of labeled images for each category, and can only classify objects into categories that have been seen. As recognition tasks evolve towards large-scale and fine-grained categories, it is difficult to meet these requirements. For example, many object classes, such as critically endangered birds and rare plant species, often follow a long-tailed distribution~\cite{zhu2014capturing} and we can not easily collect their images beforehand. Moreover, fine-grained annotation of a large number of images is laborious and even requires annotators with specialized domain knowledge~\cite{lampert2009learning, wah2011caltech, xiao2010sun}. These challenges motivate the rise of zero-shot recognition (ZSR) algorithms, in which many classes have no labeled images~\cite{palatucci2009zero, lampert2014attribute}.

\begin{figure}
  \centering
  \includegraphics[width=0.43\textwidth]{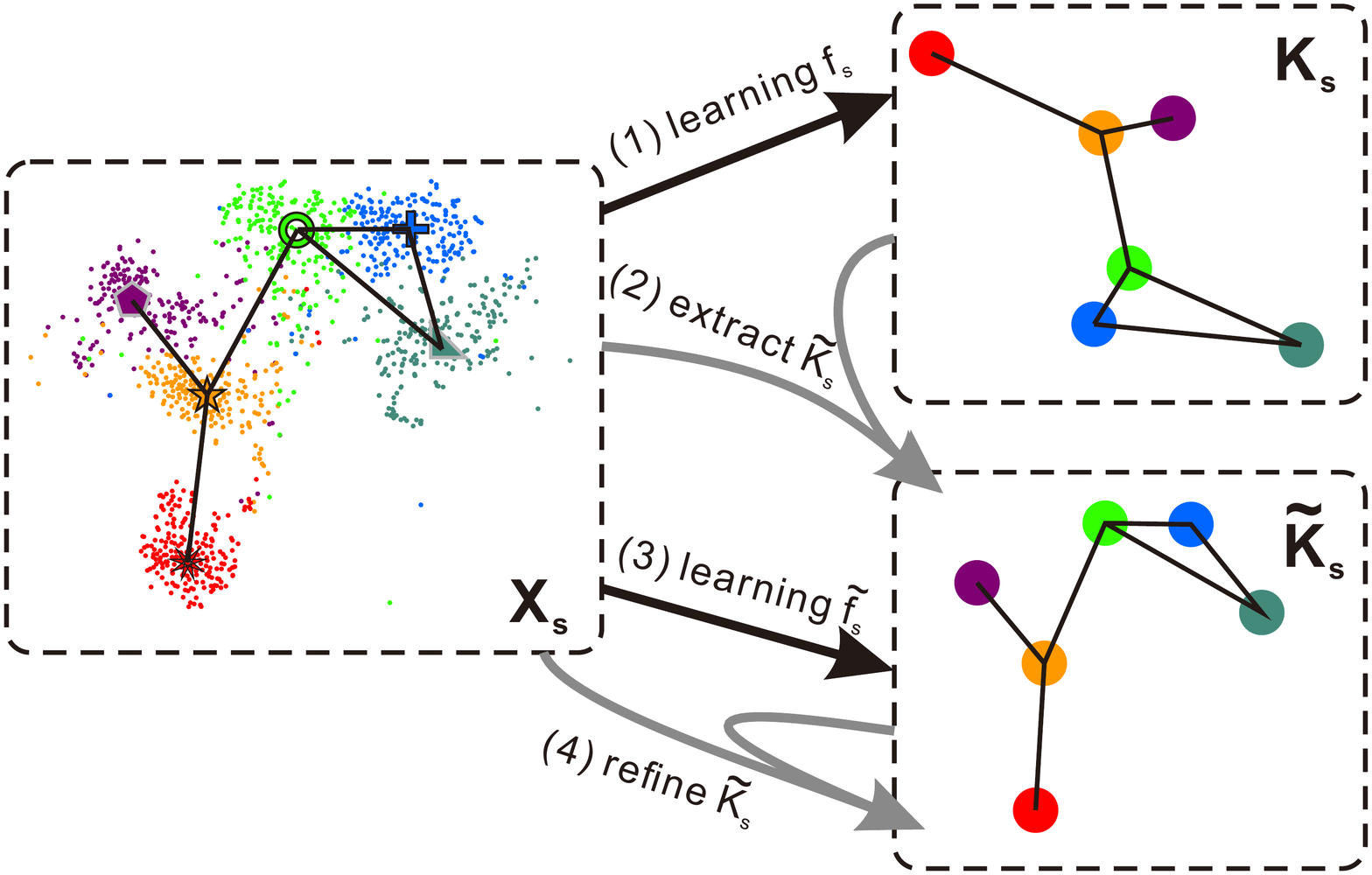}\\
  \caption{Illustration of our proposed method for ZSR. All object classes present two different class-level manifolds in $\XM_s$ and $\KM_s$ respectively, as shown in the subgraphs. Two parallel paths, starting with the same space $\XM_s$, arriving at different spaces $\KM_s$ and $\tilde{\KM}_s$, represent two visual-semantic mappings $f_s$ and $\tilde{f}_s$. Current ZSR methods only need a single path, \ie $f_s$, to project $\XM_s$ onto $\KM_s$, and predict labels in $\KM_s$. Our method uses dual paths setup and includes three steps: (1) learning $f_s$ from two heterogeneous space $\XM_s$ and $\KM_s$; (2) extract underlying class-level manifold in $\XM_s$ and generate $\tilde{\KM}_s$ that is homologous to $\XM_s$; (3) iteratively align two manifolds in $\XM_s$ and $\tilde{\KM}_s$ to obtain $\tilde{f}_s$ and refined $\tilde{\KM}_s$.
  }
  \label{fig:zsl}
\end{figure}

Current ZSR algorithms widely adopt an effective methodology of introducing some intermediate semantic embedding space $\KM$ between input image feature space $\XM$ and output label space $\LM$. The space $\KM$ contains a number of \emph{semantic embeddings} (abbreviated as embedding), which can be attribute vectors that have been manually defined~\cite{yu2013designing, akata2013label, lampert2014attribute, romera2015embarrassingly, huang2015learning, hwang2014unified}, or word vectors that have been automatically extracted from auxiliary text corpus~\cite{norouzi2013zero, akata2015evaluation, frome2013devise,mikolov2013distributed,pennington2014glove}.
Being a more semantic counterpart to object labels, that is, each attribute vector or word vector corresponds to a unique object class, the embeddings can establish the inter-class connections. For example, the attributes \eg \emph{furry}, \emph{striped} and \emph{four-legged} \etc, are shared among all categories and can be transferred to predict the unseen tigers from the seen zebras, cats and so on.

Compared with the class labels, the embeddings own several special properties. (1) They present a more complicated geometric structure in the space $\KM$ than an ordinary one of typical label representations, \eg one-hot vectors in the space $\LM$, which are distributed on the vertices of hypersimplex with same edge length. This extraordinary geometric structure, namely \emph{semantic manifold} in this paper, can encode the relationship between seen and unseen classes, which is missing in the label space $\LM$. (2) Different embeddings have their own characteristic manifold structures, which can lead to obvious variation in recognition performance. For example, on the same dataset AwA~\cite{lampert2009learning}, attribute vectors usually achieve better recognition performance on unseen classes than word vectors~\cite{akata2015evaluation, wang2016relational}. (3) The embeddings need to be constructed in advance and remain constant during the learning period.

These properties naturally raise several issues worthy of further study. First, \emph{what kind of semantic manifold in $\KM$ can be used for ZSR?}~\cite{romera2015embarrassingly} has demonstrated that a $\KM$ consisted of orthogonal or random vectors is failed in ZSR, but more discussion on this issue deserves to be expected. Second, \emph{why does the ZSR performance change with different $\KM$?} It seems that the manifold structure in $\KM$ is one of the key factors causing this variance, but the intrinsic connection between them is still lack of in-depth analysis. Third, \emph{how to construct a better $\KM$ to enhance the recognition performance on unseen classes?} Some work has yielded encouraging results.~\cite{akata2015evaluation} proposed to learn task-oriented word vectors for Dogs dataset from a specialized collection of corpora.~\cite{reed2016learning} proposed a deep learning framework to learn new embedding through the joint training of image and text data.
Both strategies proved to be feasible by experimental results, but they need to collect a lot of side information to help the training.
In contrast,~\cite{zhang2015classifying} learned new latent embedding from a given $\KM$ by supervised dictionary learning.
It is worth noting that all of these methods do not consider using the underlying manifold information in $\XM$ to construct $\KM$,
which makes it not correlated to $\XM$.

In this paper, we focus on addressing above key problems with the ideas from manifold alignment~\cite{wang2008manifold, wang2011heterogeneous}.
Similar to $\KM$, $\XM$ also contains an intrinsic manifold structure, especially for deep features.
In ZSR, we need to align two different manifolds in $\XM$ and $\KM$ by learning a visual-semantic mapping $f_s$ on seen classes. Directly learning such mapping is a very challenging task, thus we propose to transform it as a joint optimization problem of $\KM$ and $f_s$, which results in surprising results even with simple linear $f_s$.

In our work, we first answer what kind of semantic manifold in $\KM$ can provide a useful intrinsic relationship between seen and unseen classes for ZSR.
Then, we propose a measure of \emph{inter-class semantic consistency} for evaluating the matching degree between two semantic manifolds.
In particular, based on this measure, we derive an important conclusion, which announces a connection between the semantic manifold and the transfer ability of $f_s$ on unseen classes.
That means, the more the two manifolds in $\XM$ and $\KM$ are consistent, the better the mapping $f_s$ can align them and the higher the recognition accuracy can be achieved on unseen classes.
Motivated by this conclusion, we propose a effective learning strategy for solving ZSR problem, which alternately optimizes the mapping $f_s$ and the semantic space $\KM$, and gradually make the semantic manifold in $\KM$ more consistent with that in $\XM$.

To summarize, our main contributions are as follows.
\begin{itemize}
  \item We formalize the intrinsic relationship between the semantic manifolds and the transfer ability of the visual-semantic mapping $f_s$, which reveals the importance of optimizing semantic manifold in the development of new ZSR algorithms.
  \item We introduce a novel idea to cast ZSR problem as joint optimization of the manifold structure in the semantic space $\KM$ and the visual-semantic mapping $f_s$. Benefit from this idea, we can compensate for the lack of the transfer ability of $f_s$ by refining the manifold structure in $\KM$, especially when two manifolds in $\XM$ and $\KM$ are seriously inconsistent.
  \item We propose a new framework, namely dual visual-semantic mapping paths (DMaP), to solve this joint optimization problem. Our algorithm can learn not only a optimized visual-semantic mapping $f_s$ but also a new semantic space which is correlated to $\XM$. Our experiments show that using this optimized semantic space can significantly enhance the transfer ability of $f_s$ on unseen classes.
  \item We test our approach on four datasets: Animals with Attributes, Caltech-USCD Birds \cite{wah2011caltech}, Standford Dogs \cite{dataset2011novel} and ImageNet, and evaluate it on two different ZSR tasks: conventional setup and generalized setup (See details in next section). Our results in both tasks have achieved state-of-art performance.
\end{itemize}


\section{Related work}
\label{sec:related}

We focus on the following three aspects to compare our proposed approach and related work.

\vskip 0.2 cm

\noindent \textbf{Visual-semantic mapping path.}\quad
From visual-semantic connection point of view, all ZSR methods need to construct a mapping path from the image feature space $\XM$ to the semantic space $\KM$. Some methods directly project $\XM$ into $\KM$ by learning a visual-semantic mapping $f_s$~\cite{lampert2009learning, jayaraman2014decorrelating, lampert2014attribute, romera2015embarrassingly, jayaraman2014decorrelating, akata2015evaluation, xian2016latent}, while others indirectly achieve the same purpose through introducing the intermediary spaces. For example,~\cite{gan2016learning} proposes to transform $\XM$ to a new feature space first by using a kernel projection, then this new feature can be readily used in the learning of $f_s$.~\cite{fu2015transductive} suggests to project $\XM$ and $\KM$ into a shared embedding space simultaneously, then $f_s$ is learned in new space via CCA.~\cite{zhang2015classifying} proposes to separately project $\XM$ and $\KM$ into two new sparse coefficient spaces based on dictionary learning, then $f_s$ can be learned to connect two new spaces.
All these works need to learn a projection $f_s$ to align two manifolds which originate from two uncorrelated spaces $\XM$ and $\KM$, respectively.
Since $\XM$ and $\KM$ are heterogeneous, \eg one is image feature space and another is textual semantic space, mandatory training of $f_s$ will expose it to the risk of increased complexity and over-fitting on seen classes. Our approach uses a different strategy which creates two parallel visual-semantic mapping paths, and the semantic manifold can be transferred from one path to another for generating new semantic space, as shown in Fig.~\ref{fig:zsl}. Benefit from this transfer mechanism, a new visual-semantic mapping between two homogeneous spaces is learned, which can obtain better transfer ability.

\vskip 0.2 cm

\noindent \textbf{A taxonomy of ZSR methods.}\quad
Based on the usage of image data of unseen classes during testing,
we classify the ZSR works into two categories, namely the inductive ZSR and transductive ZSR.
(1) \emph{Inductive ZSR}: Most ZSR works are considered to be inductive, which receive the unseen samples serially during testing, and are the most direct and intuitive methods~\cite{lampert2014attribute, jayaraman2014zero, fu2015zero, zhang2015classifying, wang2016relational}.
(2) \emph{Transductive ZSR}: Due to the manifold structural information exists in unseen samples, transductive ZSR works process them in parallel and make use of the underlying manifold information to boost ZSR performance~\cite{rohrbach2013transfer, fu2015transductive, kodirov2015unsupervised}. For example, the graph-based label propagation strategy is widely used in transductive ZSR.
Our approach employs the transductive ZSR setting and use a simple transductive learning strategy: averaging the k-nearest neighbours to exploit the manifold structure of the test data.

\vskip 0.2 cm

\noindent \textbf{More generalized ZSR settings.}\quad
Current ZSR works are evaluated on default setting that assumes the absence of seen classes during testing, thus we only need to discriminate among unseen categories~\cite{lampert2014attribute}.
In~\cite{chao2016empirical}, they advocate a new generalized zero-shot recognition (gZSR) setting, where test data are from both seen and unseen classes and we need to classify them into whole label space.
In this paper, we also test our method on gZSR setting and the experiments demonstrate the effectiveness.

\section{Methodology}
\label{sec:method}
\subsection{Problem Setting}
Let $\LM_s=\{l_s^1, ..., l_s^k\}$ denotes a set of $k$ seen class labels and $\LM_u=\{l_u^1, ..., l_u^l\}$ a set of $l$ unseen class labels with $\LM_s \cap \LM_u = \varnothing$. In $p$-dimensional semantic embedding space $\KM$, their corresponding embedding are $\KM_s = \{\kk_s^1, ..., \kk_s^k\}$ and $\KM_u = \{\kk_u^1,..., \kk_u^l\}$. Suppose we have a labeled training dataset $\DM_s = \{\x_i, \kk_i, y_i\}_{i=1}^{n}$ of $n$ samples, where $\x_i \in \XM_s=\{\x_1, ..., \x_{n}\}$ is the feature representation of image $i$, $\kk_i \in \KM_s$ and $y_i \in \LM_s$.
Given a new testing data $\x_j$, the problem of ZSR is thus to estimate its semantic embedding $\kk_j$ and the label $y_j$.
Typical ZSR methods take a two-stage approach: (1) predicting the embedding $\kk_j$ by a learned visual-semantic mapping $f_s: \XM_s \rightarrow \KM_s$; (2) inferring class label by comparing $\kk_j$ to the embedding of either $\KM_u$ in default ZSR setting, or $\KM_s \cup \KM_u$ in gZSR setting.

\subsection{Pre-Inspection of Semantic Space $\KM$}
For a given embedding, \eg attribute vectors, word vectors or their concatenations, we usually use them in our models directly and assume their effectiveness of transferring $f_s$ from seen to unseen classes.
However, for different partitions of seen and unseen classes, their semantic manifolds may have natural defects for some ZSR methods that can cause ZSR task to fail.
Here, we suggest a proposition to detect this manifold defect.

\begin{figure}
  \centering
  \includegraphics[width=0.43\textwidth]{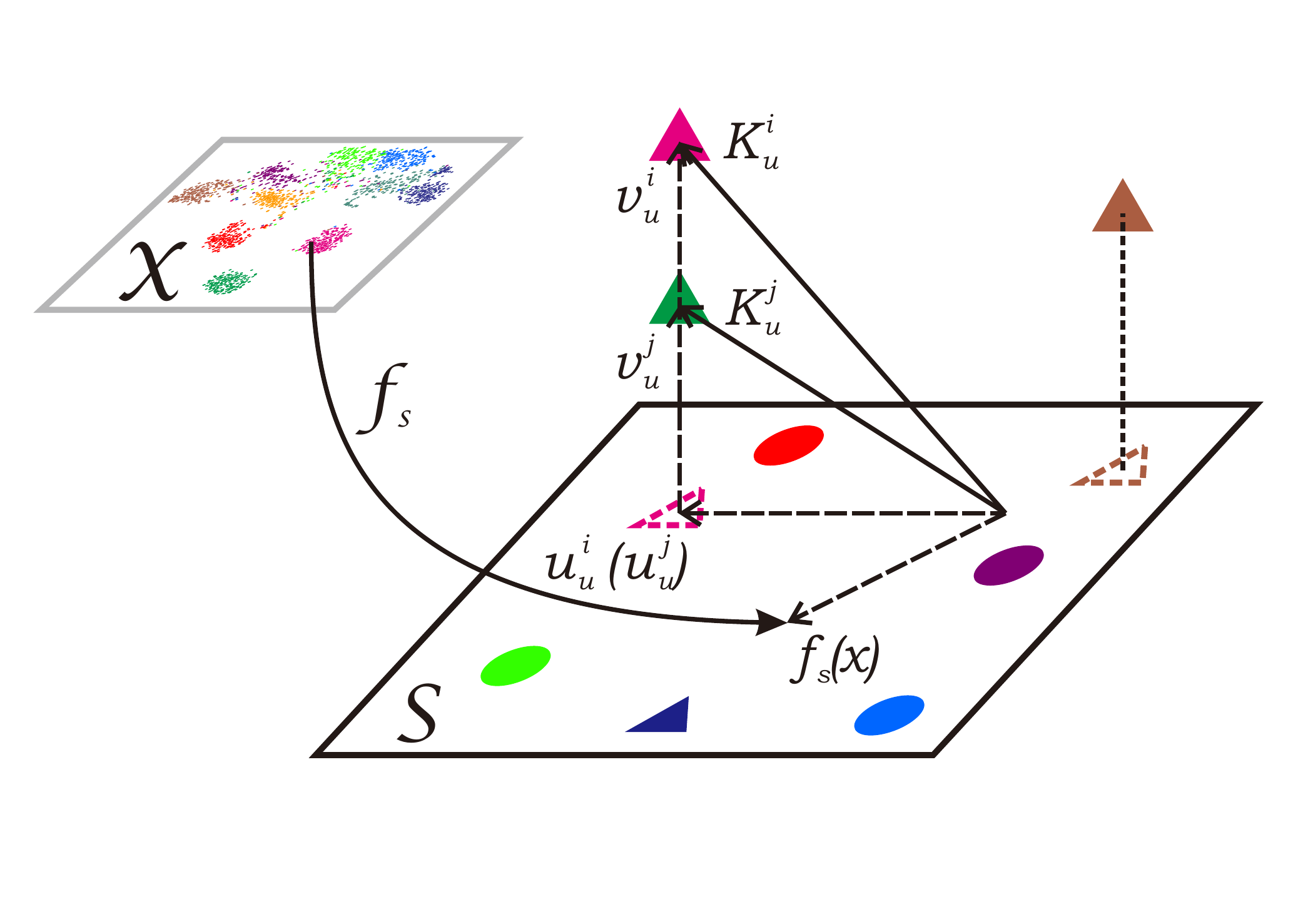}\\
  \caption{Illustration of the proposition. The circles and the triangles denote seen and unseen classes, respectively. $\SM$ denotes the subspace spanned by $\KM_s$. We show the orthogonal projections of unseen classes onto the $\SM$ in dashed triangles. More information please refer to the text.}
  \label{fig:proposition}
\end{figure}

\textbf{Proposition.} \emph{For the embedding of two unseen classes in the semantic space $\KM$, if their orthogonal projections onto the subspace $\SM$ spanned by the embedding of seen classes are equal, then $\KM$ has no transfer ability of these two unseen classes for ZSR.}

\emph{Proof.} As is shown in Fig.~\ref{fig:proposition}, suppose $\SM$ is the subspace spanned by $\KM_s$, i.e. $\SM = span(\KM_s)$. $\forall \kk_u^i \in \KM_u$, let $\u_u^i \in \SM $ be its orthogonal projection onto $\SM$, i.e. $\u_u^i = \K_s \alp_i,
s.t. \, \alp_i = \arg\min_{\alp_i} ||\kk_u^i - \K_s \alp_i||$, we have $\kk_u^i = \u_u^i +\vv_u^i$, where $\vv_u^i \perp \SM$ and $\K_s = [\kk_s^1, ..., \kk_s^k]$.
Given a test image $\x$ and its embedding $f_s(\x)$, we have $\langle f_s(\x), \kk_u^i \rangle=\langle f_s(\x), (\u_u^i+\vv_u^i) \rangle = f_s(\x)^T \u_u^i$. Likewise,
$\forall \kk_u^j \neq \kk_u^i$, we have $\langle f_s(\x), \kk_u^j \rangle = f_s(\x)^T \u_u^j$. If $\u_u^i = \u_u^j$, then $\langle f_s(\x), \kk_u^i \rangle  = \langle f_s(\x), \kk_u^j \rangle$.
Thus, these two unseen classes can not be distinguished.

The manifold defect in $\KM$ can be observed when the number of seen classes is much smaller than the number of unseen classes.
Hence, this proposition is desirable for such scenarios and can be considered as a pre-inspection step before implementing ZSR.
In addition, $\alp_i$ in the proposition defines an important inter-class relationship between seen and unseen classes, that will be used in the next subsection.

\subsection{Inter-class Relationship Consistency}
\label{sec:irc}
As reported in many ZSR works, using the same model and $\XM$, different $\KM$ could cause obvious variation in recognition performance.
For example, when predicting unseen animals in AwA dataset, manually annotated attributes usually achieve better performance than word vectors.
Intuitively, we believe that the attributes are more abstract and semantic than word vectors.
However, further experimental results show that, using the same model and $\KM$, different $\XM$ could also cause changes in recognition performance.
Thus, it is natural to infer that some association between $\XM$ and $\KM$ is the key to recognition performance.

In order to clearly understand this connection, we try to provide a formalized explanation from the view of semantic manifold consistency.
We first assume that there is an underlying class-level manifold in the image feature space $\XM$, which is more abstract than the manifold at instance level in the same space.
This class-level manifold is composed of abstract class prototypes or exemplars extracted from the instance-level manifold, as shown in Fig.\ref{fig:zsl}.
We denote the $k$ seen class prototypes and $l$ unseen class prototypes as $\tilde{\X}_s =[\tilde{\x}_s^1,...,\tilde{\x}_s^k]\in \RB^{d\times k}$ and $\tilde{\X}_u =[\tilde{\x}_u^1,...,\tilde{\x}_u^l] \in \RB^{d \times l}$, respectively.
In accordance with the above proposition, we extract the inter-class relationship matrix $\R_x = [\alp_1, ..., \alp_l] \in \RB^{k\times l}$ in $\XM$ as follows:
\begin{equation}
    \alp_i = \arg\min_{\alp_i} ||\tilde{\x}_u^i - \tilde{\X}_s \alp_i||_2 + \lambda \Omega(\alp_i),
\label{eq:relat}
 \end{equation}
where $\tilde{\x}_u^i$ is the prototype of $i$-th unseen class and $\alp_i$ denotes its association with seen classes.  $\lambda$ is the trade-off parameter and $\Omega(\alp_i)$ is a regularizer on $\alp_i$.
Similarly, we can extract the inter-class relationship matrix $\R_k = [\bet_1, ..., \bet_l]\in \RB^{k \times l}$ in $\KM$ in the same way.

\textbf{Inter-class Relationship Consistency.} If $\tilde{\X}_s\R_x = \tilde{\X}_s \R_k$, then we claim that two semantic manifolds in $\XM$ and $\KM$ have consistent inter-class relationship, or \emph{inter-class relationship consistency} (IRC).

For $\kk_u^i \in \KM_u$, let $\u_u^i$ be its orthogonal projection onto $span(\KM_s)$, i.e. $\u_u^i = \K_s \alp_i$.
If we have learned a linear visual-semantic mapping $f_s$, projecting $\tilde{\x}_s^i$ to $\kk_s^i$, and IRC is satisfied, then we derive a nice conclusion for ZSR, as shown next.

\textbf{Corollary 3.2. } \emph{If two semantic manifolds in $\XM$ and $\KM$ have consistent inter-class relationship, then $\forall i \in [1,...,l],f_s(\tilde{\x}_u^i) = \u_u^i$.}

\emph{Proof.} If IRC is satisfied, then $\tilde{\X}_s\alp_i = \tilde{\X}_s \bet_i$ for $i$-th unseen class. According to the homomorphism of linear mapping, for $\tilde{\x}_u^i$, we have $f_s(\tilde{\x}_u^i) = f_s(\tilde{\X}_s \alp_i) = f_s(\tilde{\X}_s \bet_i) = \K_s\bet_i = \u_u^i$.

From the proposition, we have known that $\u_u^i$ and $\kk_u^i$ are in one-to-one correspondence, therefor $\tilde{\x}_u^i$ is able to associate a unique $\kk_u^i$ via $f_s(\tilde{\x}_u^i)$ . In other words, IRC can ensure the transfer ability of $f_s$ from seen to unseen classes.
However, IRC is often violated in real circumstances if $\tilde{\X}_s\R_x \neq \tilde{\X}_s \R_k$, \eg $\XM$ and $\KM$ are heterogeneous that they have inherently inconsistent inter-class relationship.

\textbf{Consistency Measure. } To quantitatively evaluate inter-class relationship consistency, we provide a consistency measure,
\begin{equation}
CM(\XM|\KM) = \frac{1}{l}\sum_{i=1}^{l}
\exp(\frac{-||\tilde{\X}_s \alp_i - \tilde{\X}_s \bet_i ||_2}{||\tilde{\X}_s \alp_i ||_2 ||\tilde{\X}_s \bet_i ||_2}),
\label{eq:cm}
\end{equation}
where $||\cdot||_2$ denotes the $\ell_2$ norm. At the simplest level, we can use the mean vector of each class as the class prototype or exemplar and then compute $CM$.

\subsection{Transductive Method for ZSR}
The IRC gives us a hint that, given the image feature space $\XM$, a more semantically consistent $\KM$ can enhance the transfer ability of $f_s$. This inspires us to construct new space $\KM$ which has more consistent semantic manifold with $\XM$.
As described above, the intrinsic class-level manifold in $\XM$ can be considered as an off-the-shelf option.
To achieve this goal, we propose a simple method to jointly optimize the manifold structure in $\KM$ and the visual-semantic mapping $f_s$, during which a new homogeneous $\tilde{\KM}$ with $\XM$ is generated.

\subsubsection{Training Phrase}
We propose a three-step training process, as shown in Alg.\ref{alg:frame}, to generate new $\tilde{\KM}_s$, which is able to capture the class-level manifold in $\XM_s$. First, we learn $f_s: \XM_s \rightarrow \KM_s$ from training dataset to help infer the underlying manifold in $\XM_s$. Then, we construct new $\tilde{\KM}_s$ by means of the local manifold of $f_s$ in $\KM_s$. Finally, we alternately optimize $\tilde{f}_s: \XM_s \rightarrow \tilde{\KM}_s$ and refine $\tilde{\KM}_s$ to be more semantically consistent with $\XM_s$.

\textbf{Step 1: Learn the visual-semantic mapping. }
Without loss of generality, assume there is a linear map  $f_s: \XM_s \rightarrow \KM_s$ from image features to the embedding. Given $n$ labelled training data $\X \in \RB^{d \times n}$ and their corresponding embedding $\K\in \RB^{k \times n}$, we follow the conventional idea to learn $f_s$ by the following function,
\begin{equation}
  \arg\min_{\W} l(\W \X, \K) +\gamma g(\W),
  \label{eq:visem}
\end{equation}
where $\W$ is the parameter matrix and $g(.)$ is a regularizer. $l(.)$ is the general loss function, \eg hinge loss, logistic loss \etc. In our experiments, there is no substantial performance difference among them. In this paper, we apply the simple squared loss in Eq.~\ref{eq:visem}, which is a standard least squares problem and have a closed form solution~\cite{romera2015embarrassingly}.

\textbf{Step 2: Extract class-level manifold in $\XM_s$ and construct $\tilde{\KM}_s$. }
We aim to extract the class-level manifold in $\XM_s$ by means of the manifold in $f_s(\X)$, instead of using the mean vector of each class in $\XM_s$ mainly for two reasons. First, considering the case where instances in a class are distributed over a complex manifold, \eg crescent manifold, clearly its mean vector cannot serve as the prototype or exemplar of this class. Second, when applying this step to the testing phrase in which instances are given unlabelled, we cannot tell which instances belong to a specific category exactly, thus fail in getting their mean vector.

We exploit the idea in manifold learning that if the semantic representations of some instances and a class embedding are on the same local manifold structure, they are most likely from the same class. To be specific, for each class embedding $\kk_s^i$ , we search for its $m$ nearest neighbors in $\KM$ from $f_s(\X)$, then regard the average of those images as the class-level prototype, \ie $\tilde{\kk}_s^i$. Comparing with $\KM_s$, the new $\tilde{\KM}_s = \{\tilde{\kk}_s^i\}_{i=1}^k$ is more semantically consistent with $\XM_s$.

\textbf{Step 3: Align manifolds iteratively. }
$\tilde{\KM}_s$ captures the latent class-level manifold in $\XM_s$ and can be further refined. We alternate between (3a) learn $\tilde{f}_s: \XM_s \rightarrow \tilde{\KM}_s$ and (3b) refine $\tilde{\KM}_s$, which are learned in the same way above, until the optimization procedure converges or the maximal iteration number is reached. In practice, the algorithm can converge on the first few iterations.

\begin{algorithm}[thp]
\caption{Training algorithm of our method}
\begin{algorithmic}[1]
\STATE \textbf{Input}: Labelled training dataset $\DM_s = \{\x_i, \kk_i, y_i\}_{i=1}^{n}$, semantic embedding $\KM_s$.
\STATE \textbf{Output}: $f_s$, $\tilde{f}_s$ and $\tilde{\KM}_s = \{\tilde{\kk}_s^i\}_{i=1}^k$.
\STATE \emph{Step 1}:
\STATE Learn $f_s: \XM_s \rightarrow \KM_s$ on $\DM_s$  by Eq.~\ref{eq:visem}.
\STATE \emph{Step 2}:
\FOR {$\forall \kk_s^i \in \KM_s$}
\STATE  Find its $m$ nearest neighbors from all predictions $\{f_s(\x_i)\}_{i=1}^{n}$ and denote the corresponding images as $NN_{\KM}^m(\kk_s^i)$.
\STATE Construct new semantic embedding $\tilde{\kk}_s^i$ as the average $\frac{1}{m}\sum NN^m_{\XM}(\kk_s^i)$.
\ENDFOR
\STATE \emph{Step 3}:
\REPEAT
\STATE Learn $\tilde{f}_s: \XM_s \rightarrow \tilde{\KM}_s$.
\STATE Refine $\tilde{\KM}_s$ as formulated above.
\UNTIL{Done}
\end{algorithmic}
\label{alg:frame}
\end{algorithm}

\subsubsection{Testing Phrase}
During testing, we take $f_s$, $\tilde{f}_s$, $\tilde{\KM}_s$, $\KM_s$ and $\KM_u$ as inputs.
Given $n_t$ testing instances $\X_u \in \RB^{d\times n_t}$, we first predict their semantic representations as $f_s(\X_u)$, then we construct the jump-start $\tilde{\KM}_u$ transductively as in Step 2. Finally, for each testing  instance $\x_j$, we compare $\tilde{f}_s(\x_j)$ with new label embedding using the inner product measure $d$ and label it as the nearest class, \ie $ y_j = \arg\max_{c} d(\tilde{f}_s(\x_j), \tilde{\kk}_c)$, where $\tilde{\kk}_c \in \tilde{\KM}_u$ in ZSR and $\tilde{\kk}_c \in \{\tilde{\KM}_s \cup \tilde{\KM}_u\}$ in gZSR.

\section{Experiments}
\label{sec:experiment}

\subsection{Experimental Setup}
\textbf{Datasets } We evaluate on three small-scale benchmark datasets and a large-scale dataset in our experiments: the Animals with Attributes (\textbf{AwA})~\cite{lampert2009learning},  Caltech-UCSD Birds-200-2011 (\textbf{CUB})~\cite{wah2011caltech}, Standford Dogs (\textbf{Dogs})~\cite{dataset2011novel} and ImageNet ILSVRC 2012 (\textbf{ImageNet})~\cite{russakovsky2015imagenet}. \textbf{AwA} consists of 30,475 images of 50 image classes, each containing at least 92 images, paired with a human provided 85-attribute inventory and corresponding class-attribute associations. We follow the commonly agreed experimental protocol in the literature, \ie 40 classes for training and 10 for testing. \textbf{CUB} is a fine-grained dataset with 312 attributes annotated for 200 different bird classes. It contains 11,788 images in total. Following~\cite{akata2015evaluation}, we use the same zero-shot split with 150 classes for training and 50 for testing. \textbf{Dogs} contains 19,501 images of 113 fine-grained dog species, with no human-defined attributes annotated. 85 classes are used for training, while the rest for testing. The large-scale \textbf{ImageNet} dataset contains 1,000 categories and more than 1.2 million images. We follow the 800/200 split~\cite{frome2013devise} to perform our method.

\textbf{Choices for $\XM$ and $\KM$ } For all four datasets, we choose 3 types of deep features for $\XM$ due to their superior performance, as well as the prevalence in ZSR literature. They are extracted from VGG~\cite{simonyan2015very}, GoogLeNet~\cite{szegedy2014going} and ResNet~\cite{he2015deep} and are denoted as \emph{vgg}, \emph{goog} and \emph{res}, respectively. Compared with the low-level features, they have a richer semantic manifold. For $\KM$, we adopt 2 types of semantic embedding, \ie human annotated attributes (denoted as \emph{att}) and continuous word vector representations (Word2Vec) learned from Wikipedia. For Word2Vec, 2 types are included, \ie \emph{skipgram}~\cite{mikolov2013distributed} and \emph{glove}~\cite{pennington2014glove}.

\textbf{ZSR tasks and evaluation metrics } We consider two different ZSR settings in a variety of experiments: conventional ZSR (\textbf{cZSR}) and generalized ZSR (\textbf{gZSR}). In cZSR, we train on seen classes and test on unseen ones, where the test instances are assumed to be from the unseen categories (denoted as $\UM\rightarrow \UM$). While in gZSR, we assume the test instances to come from all the target classes (denoted as $\UM \rightarrow \TM$).
We report the average classification accuracy on unseen classes.

\textbf{Implementation details } We learn $f_s$ and $\tilde{f}_s$ using the simple linear mapping in~\cite{romera2015embarrassingly}. It is extremely easy to be implemented, requiring just one line of code for training. $f_s$ is learned by optimizing: $\arg\min_{\V}||\X_s^T \V \K_s - \Y_s||^2_F + \Omega(\V)$, where $\X_s$ and $\Y_s$ denote the training instances and training labels, respectively. We name our proposed method in inductive and transductive manners as \textbf{DMaP-I} and \textbf{DMaP-T}, respectively. And DMaP-I is to conduct classification directly after learning $f_s$. We use $\ell_2$-norm to extract the relationship $\alp_i$ and fix the parameter $\lambda$ in Eq.~\ref{eq:relat} as $10^{-4}$. And we fix a consistent number $m=100$ of nearest neighbors for all these datasets. 

\begin{table}[!htp]
\caption{ZSR average accuracy (\%) and CM values using different pairs of $\XM$ and $\KM$ by DMaP-I on CUB. $v+g+r$, $gl$ and $sk$ are short for $vgg+goog+res$, $glove$ and $skipgram$, to save space.}
\begin{tabular}{|c|c|c|c|c|c|c|}
  \hline
  \multirow{2}{*}{\diagbox{$\KM$}{$\XM$}} & \multicolumn{2}{c|}{\emph{goog}} &\multicolumn{2}{c|}{\emph{vgg+goog}}& \multicolumn{2}{c|}{\emph{v+g+r}} \\
  \cline{2-7}
   & Acc & CM & Acc & CM & Acc & CM \\
  \hline
  \emph{att} &51.09 & 0.47&	52.83&	0.57	&\textbf{54.55}&	0.63 \\
  \hline
  \emph{gl} &23.69	&0.38	&24.55&	0.48&25.72&0.55 \\
  \hline
  \emph{sk} &26.28&0.40&26.38&0.49&\textbf{27.48}&0.56\\
  \hline
\emph{att+gl} & 51.23&0.51&53.38&0.60&55.14&0.66 \\
  \hline
  \emph{att+sk} &51.62&0.52&53.48&0.61&\textbf{56.34}&0.67\\
  \hline
\end{tabular}
\label{tab:K}
\end{table}

\subsection{Validation of Inter-class Relationship Consistency}
In the first set of experiments, we verify whether different semantic embedding space $\KM$ has a different IRC with $\XM$ and test the impacts of IRC on cZSR performance using DMaP-I. In addition to the spaces listed above, we compare with another two $\KM$ spaces, \ie \emph{att+skipgram} and \emph{att+glove}, where + denotes the concatenation of two embedding. We use the mean of image features of each class as prototype to extract the inter-class semantic relationship $\alp_i$, which we use to compute CM. For demonstration, we show the results on CUB in Tab.~\ref{tab:K}.

From Tab.\ref{tab:K}, we observe that ZSR performance is positively correlated to the CM value.
This not only validates our assumption that the manifold structure in $\KM$ affects ZSR performance, but also illustrates the feasibility of the manifold alignment for ZSR.

We also find that CM($\XM|$\emph{att+skipgram}) $>$ CM($\XM|$\emph{att}) $>$  CM($\XM|$\emph{skipgram}), and this trend holds true for ZSR performance as well.
This trend for performance has appeared in the ZSR literature. This suggests that these two different semantic embedding spaces contain complementary information which should be combined for ZSR.

\subsection{Evaluation of Our Method on cZSR and gZSR}
\begin{figure}
  \centering
  \subfigure{\includegraphics[width=0.46\textwidth]{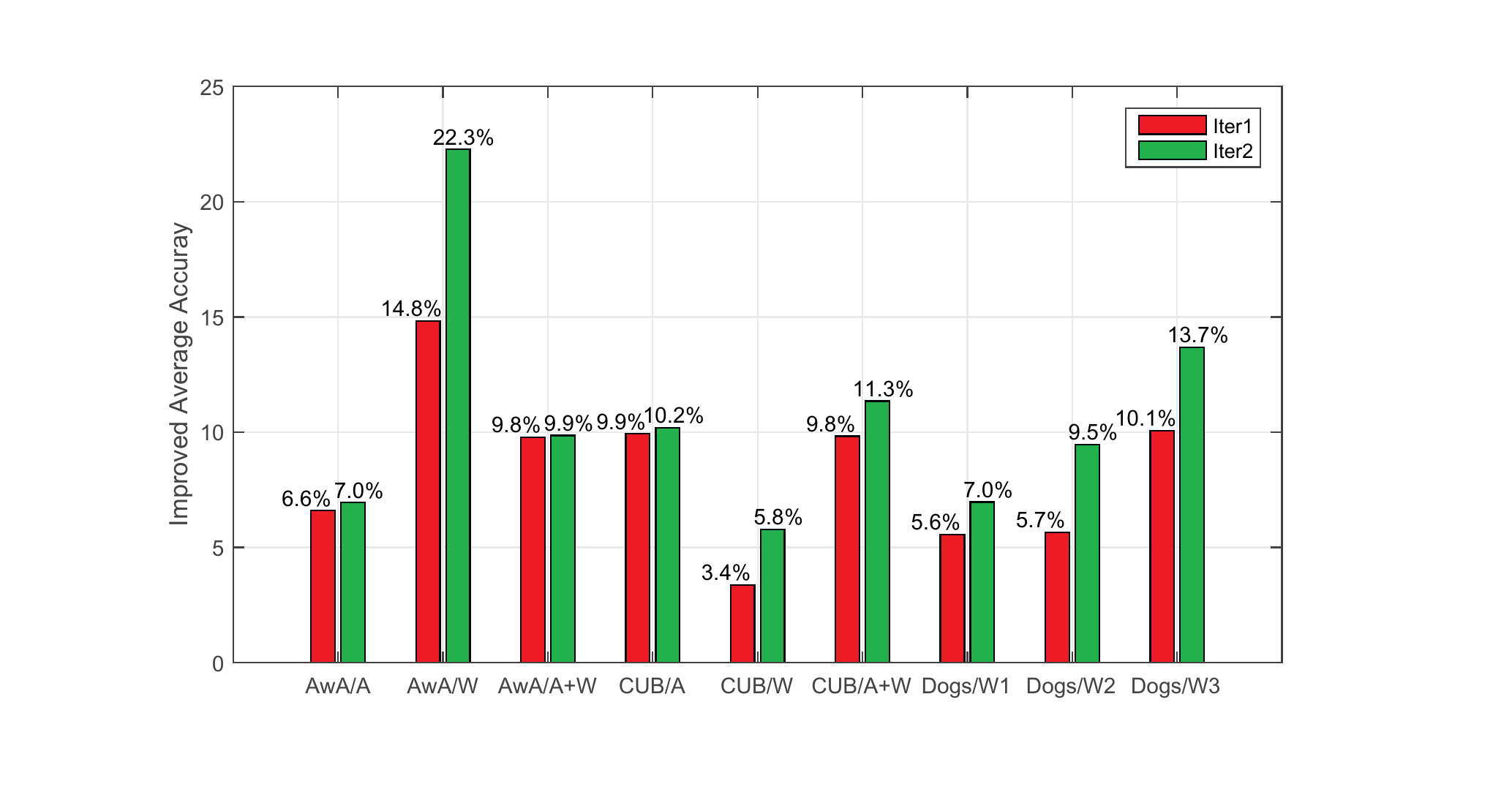}}
  \caption{Accuracy improvement using DMaP-T over DMaP-I. Results are obtained using 2 iterations.}
  \label{fig:transbar}
\end{figure}

\begin{figure}
  \centering
  \subfigure{\includegraphics[width=0.46\textwidth]{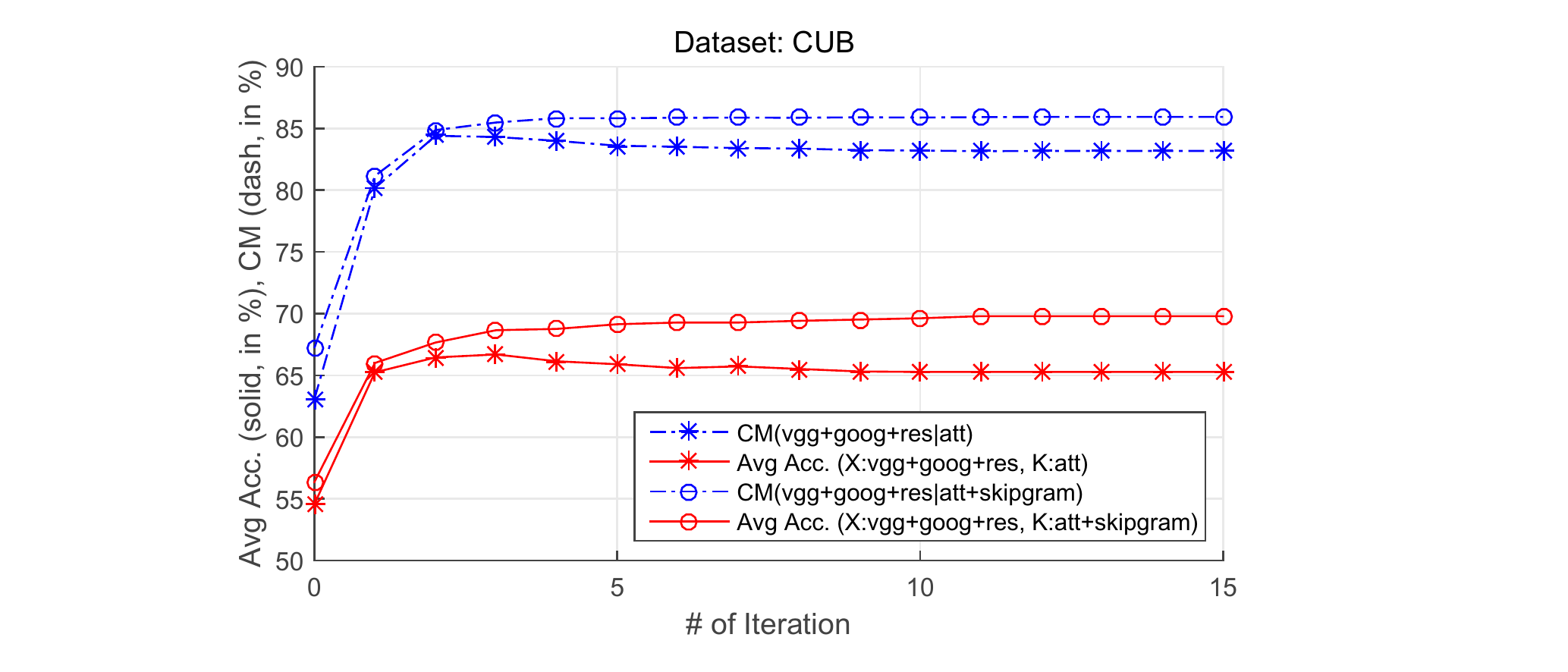}}
  \caption{ZSR average accuracy (\%) and the corresponding CM values (the y-axis) obtained with different number of iterations on CUB (the x-axis).}
  \label{fig:trans}
\end{figure}

In the second set of experiments, we evaluate our method on both cZSR and gZSR tasks. In the Step 1 of Alg.~\ref{alg:frame}, an initial mapping $f_s: \XM_s \rightarrow \KM_s$ is learned. As described in Sec.~\ref{sec:irc}, different configurations of $\XM$ and $\KM$ may result in different ZSR performance. To verify this statement, we run various configurations and show the best performance in Tab.~\ref{tab:ours}. We use \emph{att}, \emph{skipgram} and \emph{att+skipgram} for $\KM$ on AwA and CUB. While on Dogs and ImageNet, due to the lack of attributes, we use only \emph{skipgram} for $\KM$. 
Tab.\ref{tab:ours} presents the recognition accuracies of DMaP-I and DMaP-T in two iterations.

\subsubsection{Experimental results on cZSR}

\begin{table*}[!htp]
  \centering
  \caption{ZSL average accuracy (\%) achieved by our method (DMaP-I and DMaP-T with two iterations, denoted as Iter1 and Iter2 to save space) on both cZSR and gZSR tasks on AwA, CUB, Dogs and ImageNet datasets. We report top-1 accuracy on ImageNet. }
  \begin{tabular}{|c|c|c|c|c|c|c|c|}
    \hline
    \multicolumn{2}{|c|}{} & \multicolumn{3}{c|}{cZSR($\UM \rightarrow \UM$)} & \multicolumn{3}{c|}{gZSR($\UM\rightarrow \TM$)} \\
    \hline
    Dataset & $f_s: \XM \rightarrow \KM$ & DMaP-I & Iter1 & Iter2 & DMaP-I & Iter1 & Iter2 \\
    \hline
    \multirow{3}{*}{AwA} & \emph{vgg} $\rightarrow$ \emph{att}  & 78.71 &85.31  &85.66 & 17.23 & 49.66  &\textbf{52.70}\\
    \cline{2-8}
                                     &\emph{ res} $\rightarrow$ \emph{skipgram}  & 63.43 &78.25 &85.70 & 6.44 & 6.72  &18.85 \\
    \cline{2-8}
                                     &\emph{ vgg+res} $\rightarrow$ \emph{att+skipgram} &80.63 &90.42 &\textbf{90.49} & 2.72 & 10.60  & 17.82\\
    \hline
    \multirow{3}{*}{CUB} & \emph{goog} $\rightarrow$ \emph{att} &51.59 &61.52 & 61.79 &  13.55 & 24.28  & \textbf{27.83}\\
    \cline{2-8}
                                     & \emph{vgg+goog} $\rightarrow$ \emph{glove} & 24.55& 27.93 & 30.34 & 2.07 & 3.62  & 6.41\\
    \cline{2-8}
                                     &\emph{vgg+goog+res} $\rightarrow $\emph{att+skipgram}  & 56.34&66.17 &\textbf{67.69} & 7.00 &19.86  &21.86 \\
    \hline
    \multirow{3}{*}{Dogs} & \emph{vgg} $\rightarrow$ \emph{skipgram} & 26.60& 32.17&33.57 & 0.54 &2.93 &4.96\\
    \cline{2-8}
                                     & \emph{goog} $\rightarrow$ \emph{skipgram}    & 29.46&35.12 &38.92 & 0.18 &4.64  &5.10\\
    \cline{2-8}
                                     & \emph{vgg+goog}$ \rightarrow$ \emph{skipgram}    & 30.90&40.97 &\textbf{44.59} & 0.22 &4.94 &\textbf{5.10}\\
    \hline
   ImageNet & \emph{goog} $\rightarrow$ \emph{skipgram} &28.30 &38.76  &\textbf{38.94} &0.74 &12.00 &\textbf{17.00}\\
    \hline
  \end{tabular}
  \label{tab:ours}
\end{table*}

The performance improvements over DMaP-I are shown in the left three columns of Tab.~\ref{tab:ours} and Fig.~\ref{fig:transbar}. These results demonstrate that in all cases, our manifold alignment process can significantly boost DMaP-I. Using only two iterations, it can be improved by an average accuracy of 10.71\%. On AwA, the performance improvement even achieves the astonishing accuracy of 22.3\%, as shown in Fig.~\ref{fig:transbar}. And even if the initial performance of $f_s$ is relatively lower, our algorithm still has the ability to achieve good performance. In other words, even though the initial manifold in $\KM$ is of lower quality, it will still be driven to be more consistent with $\XM$. For example, on Dogs, one iteration can increase the accuracy from 30.90\% to 40.97\% impressively.

In another experiment, we test how the number of iterations affects the performance. Fig.~\ref{fig:trans} illustrates the results on CUB datasets.  On both $\KM$, a fast convergence tendency can be observed. Generally, after one or two iterations, DMaP-T can achieve remarkable improvement. Moreover, since $f_s$ is a linear mapping, the computational complexity is very low. These results once again validate the feasibility and effectiveness of our method.

\subsubsection{Experimental results on gZSR}
The right three columns of Tab.~\ref{tab:ours} summarize the accuracy on gZSR task, \ie predict testing labels from all classes. We observe that compared with results on cZSR, DMaP-I on gZSR achieves considerably poor performance, which is consistent with the phenomenon reported in~\cite{chao2016empirical}. On Dogs and ImageNet, nearly all test data from unseen classes are misclassified into the seen ones. In addition, we reproduced both DeViSE~\cite{frome2013devise} and ConSE~\cite{norouzi2013zero}, and conducted extensive ZSR experiments on ImageNet, \ie 1K for training and 21K for testing. We found that the top-1 accuracies of most classes are actually close to 0. We think Proposition 1 can give us a reasonable explanation for this phenomenon, \ie the manifold defect.
This unusual degradation in performance highlights the challenge of gZSR. However, our method can still increase the recognition accuracy significantly. On AwA, the best accuracy is $52.7\%$, which means $35.47\%$ improvement over DMaP-I, $50.3\%$ improvement over DAP and $52.3\%$ improvement over SynC~\cite{chao2016empirical}. Even on the large-scale ImageNet, we also obtain a surprising and remarkable improvement.

\begin{figure*}
  \centering
  \subfigure{\includegraphics[width=0.95\textwidth]{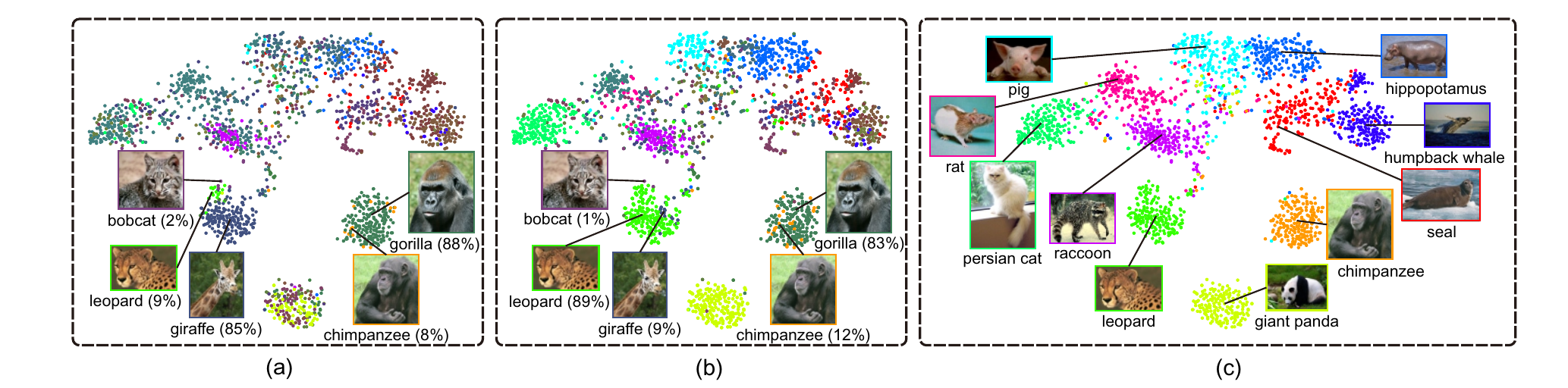}}
  \caption{Illustration of the results of $\UM \rightarrow \TM$ task on AwA dataset. (a) Results obtained by DMaP-I. (b) Results obtained by DMaP-T with one iteration. (c) Ground truth unseen class label. The percentage in parentheses denotes the proportion of the ground-truth unseen class classified as this corresponding category, \eg bobcat $2\%$ in (a) denotes $2\%$ leopard samples are inaccurately classified as bobcat. This figure is best viewed in color.}
  \label{fig:final}
\end{figure*}

\begin{figure*}
  \centering
  \subfigure{\includegraphics[width=0.95\textwidth]{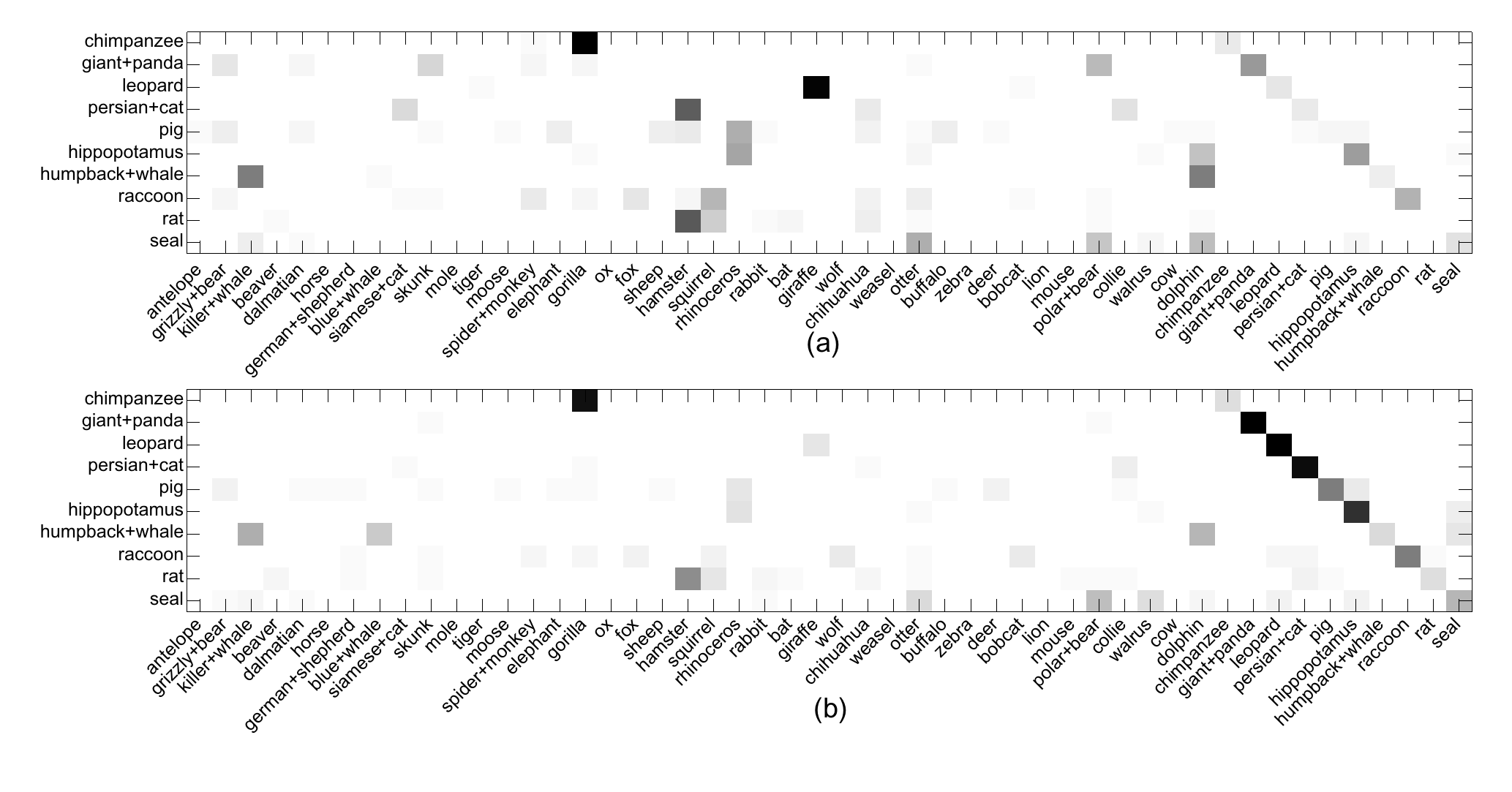}}
  \caption{Confusion matrix for recognition accuracies of $\UM\rightarrow \TM$ task evaluated on AwA dataset. (The first 40 in x-axis are seen classes, and the others are unseen ones.) (a) results obtained by DMaP-I. (b) results obtained by DMaP-T with one iteration.}
  \label{fig:confusion}
\end{figure*}

For better understanding of our method, we visualize the $\UM\rightarrow \TM$ results of each iteration using t-SNE~\cite{van2008visualizing} in Fig.~\ref{fig:final} and show the confusion matrices for DMaP-I and DMaP-T in Fig.~\ref{fig:confusion}. For clear demonstration, we only display the results on AwA. We use 40 colors with lower brightness to denote seen classes and the other 10 colors with high brightness to represent unseen ones. Instances are classified as the label shown by their color. By comparing Fig.~\ref{fig:final} (a) with Fig.~\ref{fig:final} (b), we observe that with one iteration, our method could better classify the unseen instances. For example, although ``bobcat'', ``leopard'' and ``giraffe'' have a large overlap, $89\%$ of leopard images are classified correctly after one iteration, much more than $9\%$ in DMaP-I. However, our method fails for some categories such as ``chimpanzee''. Chimpanzee images are always classified as ``gorilla''. This may be because these two classes are very close to each other on the manifold in $\XM$.

\subsection{Comparison with State-of-the-arts}
\begin{table}[!htp]
\centering
\caption{cZSR ($\UM \rightarrow \UM$) comparison on AwA, CUB and Dogs. We compare ours (achieved using 2 iteration) with the state-of-the-art results using different $\KM$, including word vector (W) and attribute (A). See Supp file for more details. `T' or `I' denotes transductive or inductive methods. `+' indicates the concatenation operation.  `--' means no result reported in the original paper. }
  \begin{tabular}{c c c c c c}
  \hline
  Methods & $\KM$ & T/I &AwA & CUB & Dogs\\
  \hline \hline
  SSE & A & I&76.23 & 30.41 & --\\
  SJE     & A/W   &I  & 66.7 & 50.1 & 33.0 \\
  SynC  & A+W & I & 72.9 & 54.7 & --\\
  LatEm & A+W  &I& 76.1& 51.7& 36.3\\
  RKT     & A+W    &I& 82.43 & 46.24& 28.29 \\
   AMP     & A+W     &I& 66 & -- & -- \\
  \hline
  TMV-HLP & A+W     &T& 80.5 & 47.9 & -- \\
   UDA & A &  T&75.6 & 40.6 & --\\
   PST& A & T & 42.7 & -- & -- \\
  \hline\hline
\multirow{3}{*}{\textbf{DMaP}} &  A & T& 85.66 &  61.79 & --\\
 & W & T& 85.70 & 30.34 & \textbf{44.59} \\
 & A+W & T& \textbf{90.49} & \textbf{67.69} & -- \\
 \hline
  \end{tabular}
  \label{tab:comp-state}
\end{table}

We provide a direct comparison between our method (denoted as \textbf{DMaP}) and three transductive ZSR methods, \ie PST~\cite{rohrbach2013transfer}, TMV-HLP~\cite{fu2015transductive} and UDA~\cite{kodirov2015unsupervised}. In addition, the performance of our approach is also compared against inductive methods, \ie AMP~\cite{fu2015zero}, SSE~\cite{zhang2015zero}, SJE~\cite{akata2015evaluation}, SynC~\cite{changpinyo2016synthesized}, LatEm~\cite{xian2016latent} and RKT~\cite{wang2016relational},  which are, to the best of our knowledge, state-of-the-art methods for ZSR. All these methods except PST use deep features to represent images in $\XM$. We report their best published results on cZSR on three benchmark datasets in Tab.~\ref{tab:comp-state}.

It is clear that our method significantly outperforms the others on all three datasets. Even if initiating from a low-quality semantic embedding space (\eg word vector representations), it can still achieve higher performance than others using  a better $A+W$. For instance, DMaP achieves the highest accuracy of 44.59\% on Dogs. In addition,~\cite{fu2015zero} reported the hit@5 accuracy on ImageNet 2010 1K is 41\%. Comparatively, on the more challenging ImageNet 2012 1K, our method achieves the remarkably 38.94\% hit@1 accuracy. This superior performance demonstrate the effectiveness of our proposed method. Note that DMaP is a very general method since the alignment process could be added to inductive DMaP flexibly. When incorporated with other inductive ZSR methods, it is expected to further improve the performance.
%

\section{Discussion and Conclusion}
\label{sec:conclusion}
We presented an analysis of the semantic embedding space for ZSR, and revealed a connection between the manifold structure and the transfer ability of visual-semantic mapping. It is reasonable to think that the inter-class semantic consistency of two spaces is the key to effective ZSR. Motivated by this, we developed a DMaP framework to generate more consistent semantic space with the image feature space as well as learn more effective visual-semantic mapping. Our method outperform the state-of-the-art approaches on four challenging datasets.

{\small
\bibliographystyle{ieee}
\bibliography{cvpr17bib}
}
%

\section*{Supplementary Material: Zero-Shot Recognition using Dual Visual-Semantic Mapping Paths}
In this supplementary material, we provide below practical details of our implementation omitted in the main text.
\section{Implementation Details}
\textbf{1. The choice of $\Omega$ in Eq.1. } During the extraction of inter-class relationship by Eq.~1 in the main text, common choice for $\Omega$ is $\ell_1$ norm or $\ell_2$ norm. When $\Omega(\alp_i) = ||\alp_i||_2$, Eq.~1 is a typical ridge regression problem and we exploit the global structure of $\X_s$ to reconstruct the inter-class relationship. When $\Omega(\alp_i) = ||\alp_i||_1$, where Eq.~1 becomes a sparse coding problem, the local structure of $\X_s$ is exploited. In our experiments, we choose $\ell_2$ norm for $\Omega$.

\textbf{2. The mapping function $f_s$. } Let us denote $n$ labelled training data from $k$ seen classes as $\X_s \in \RB^{d \times n}$ and their ground truth labels are $\Y_s \in \{-1, 1\}^{n \times k}$, each row of which contains only one positive entry indicating the class it belongs to. Also, the label embeddings of seen classes are indicated by columns of $\K_s \in \RB^{p\times k}$.  We adopt the linear mapping function in~\cite{romera2015embarrassingly} to learn the visual-semantic mapping $f_s$. The objective function in Eq.3 becomes:
  \begin{equation}
    \arg\min_{\V} ||\X_s^T \V \K_s - \Y_s||^2_F + g(\V),
  \end{equation}
  where $\V \in  \RB^{ d \times p}$ is the parameter we learn and $g(\V) = \gamma ||\V \K_s||_F^2 + \eta ||\X_s^T \V||^2_F + \gamma \eta ||\V||^2_F$. Thus its solution can be expressed in closed form:
  \begin{equation}
  \V = (\X_s \X_s^T + \gamma \I)^{-1} \X_s \Y_s\K_s^T (\K_s\K_s^T + \eta \I)^{-1}.
  \label{eq:fs}
  \end{equation}
  where $\I$ is the identity matrix.

\textbf{3. Values of hyper-parameters. } There are a few free hyper-parameters to be tuned in our approach, \ie $\lambda$ in Eq.~1 (in the main text), $\gamma$ and $\eta$ in Eq.~\ref{eq:fs}. $\lambda$ is set to $10^{-4}$. $\gamma$ and $\eta$ are chosen from range $10^{[1.2, 1.5]}$ and $10^{[4.2, 5.4]}$, respectively.

\textbf{4. Dimensions of the image features and the semantic embeddings. } We conduct experiments with deep features on all datasets, extracted by VGG~\cite{simonyan2015very}, GoogLeNet~\cite{szegedy2014going} and ResNet~\cite{he2015deep}. For VGG and ResNet, we use the 1000-dimensional activations of last fully connected layer as features, and for GoogLeNet we extract features by the 1024-dimensional activations of the top-layer pooling unites. We choose two different types of word vectors in our experiments, \ie \emph{skipgram}~\cite{mikolov2013distributed} and \emph{glove}~\cite{pennington2014glove}. They are trained on the Wikipedia corpus and their dimensions are set to 500 and 300, respectively.

\section{Additional experimental results}
We present in this section some additional experimental results on zero-shot recognition.

\subsection{Visualization of the proposed DMaP-T}
\begin{figure*}
  \centering
  \subfigure{\includegraphics[width=0.97\textwidth]{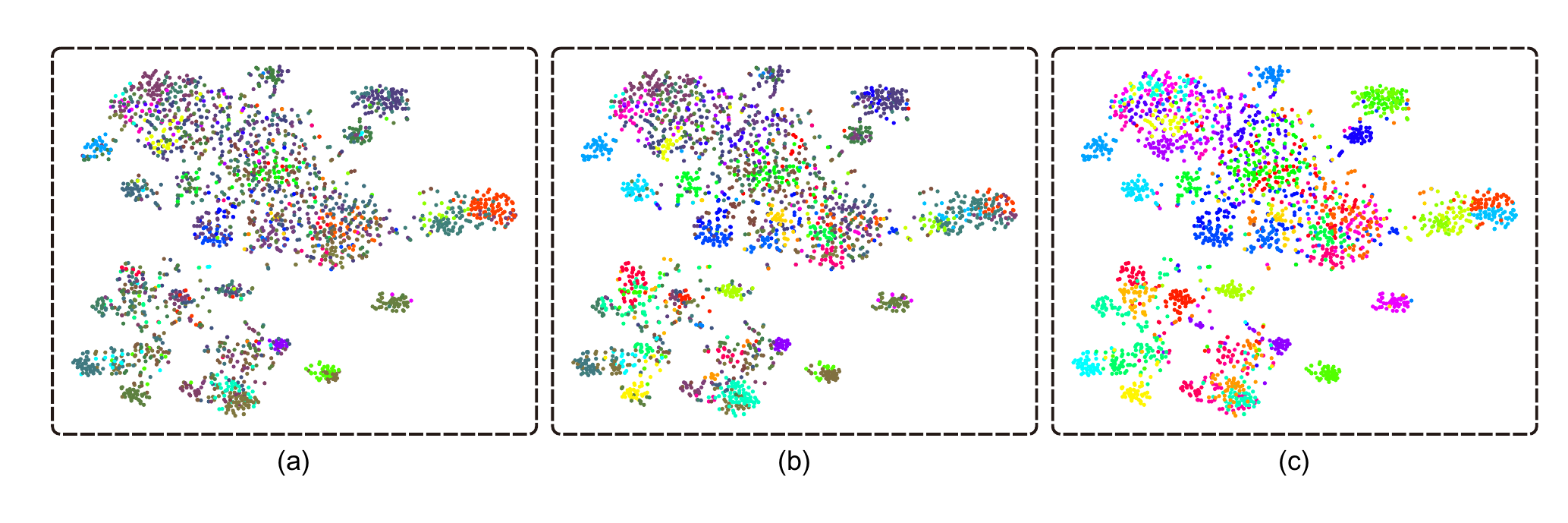}}
  \caption{Illustration of the results of $\UM \rightarrow \TM$ task on CUB dataset. (a) Results obtained by DMaP-I. (b) Results obtained by DMaP-T with one iteration. (c) Ground truth unseen class label. Dots with lower brightness denote unseen instances are mistakenly classified to the previously seen classes. The higher brightness of the whole image indicates the better recognition results. This figure is best viewed in color.}
  \label{fig:cub}
\end{figure*}

\begin{figure*}
  \centering
  \subfigure{\includegraphics[width=0.97\textwidth]{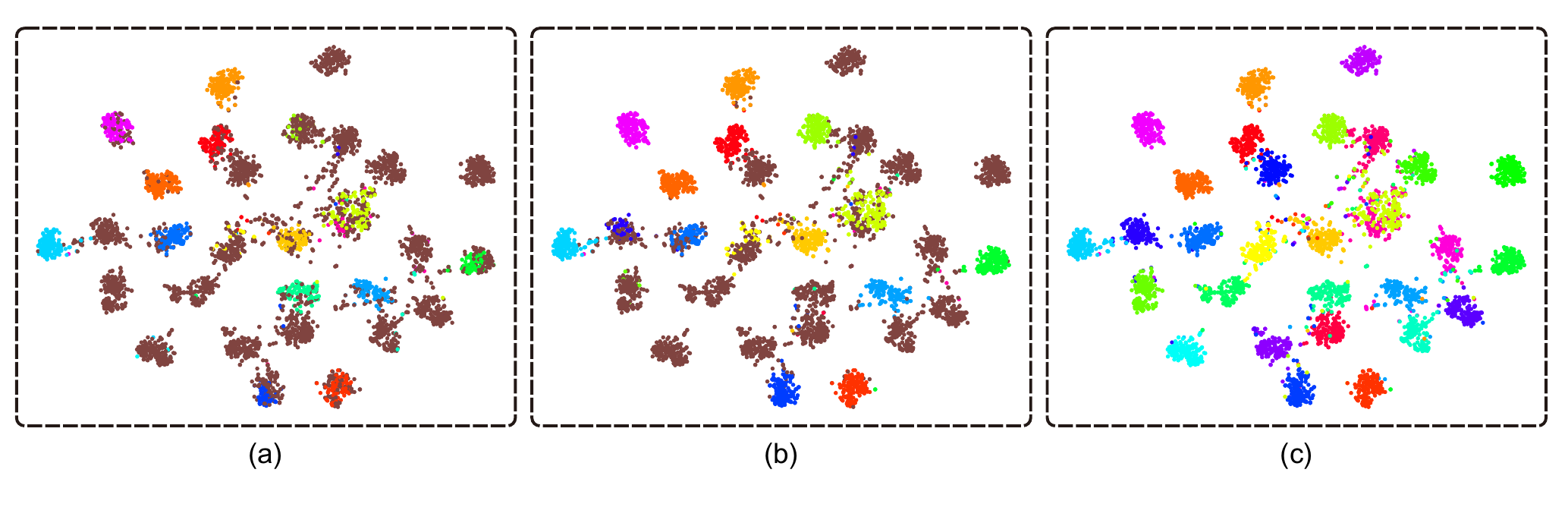}}
  \caption{Illustration of the results of $\UM \rightarrow \UM$ task on Dogs dataset. (a) Results obtained by DMaP-I. (b) Results obtained by DMaP-T with one iteration. (c) Ground truth unseen class label. The brown color dots denote unseen instances are classified to wrong classes. This figure is best viewed in color.}
  \label{fig:dogs}
\end{figure*}

In addition to Fig. 5 of the main text, we further visualize our zero-shot recognition results of $\UM\rightarrow \TM$ on CUB and $\UM \rightarrow \UM$ on Dogs in Fig.~\ref{fig:cub} and Fig.~\ref{fig:dogs}, respectively.

\subsection{Pre-inspection of Semantic Space $\KM$}
\begin{figure*}
  \centering
  \subfigure{\includegraphics[width=0.97\textwidth]{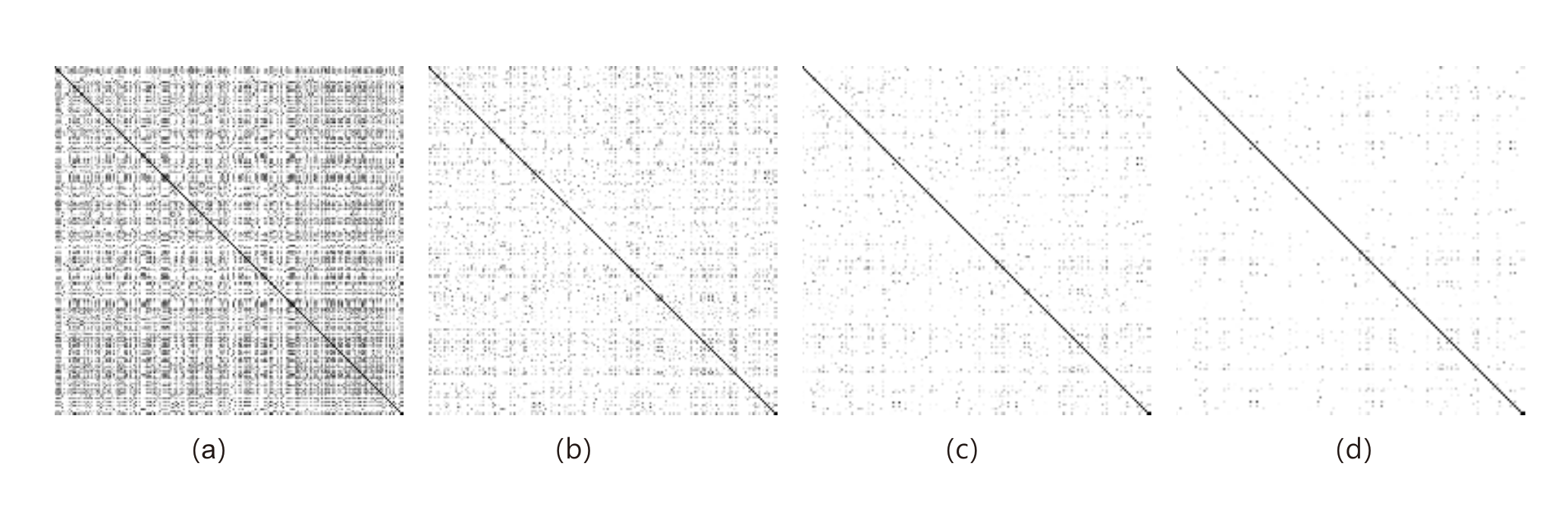}}
  \caption{Visualization of pairwise Euclidean distances among orthogonal projections of unseen classes on CUB dataset. These pairwise distances are obtained by using different seen/unseen splits. (a) Results obtained on split 10/190. (b) Results obtained on split 20/180. (c) Results obtained on split 30/170. (d) Results obtained on split 40/160.  Darker colors depicts closer distances.}
  \label{fig:cub_cm}
\end{figure*}

\begin{figure*}
  \centering
  \subfigure{\includegraphics[width=0.85\textwidth]{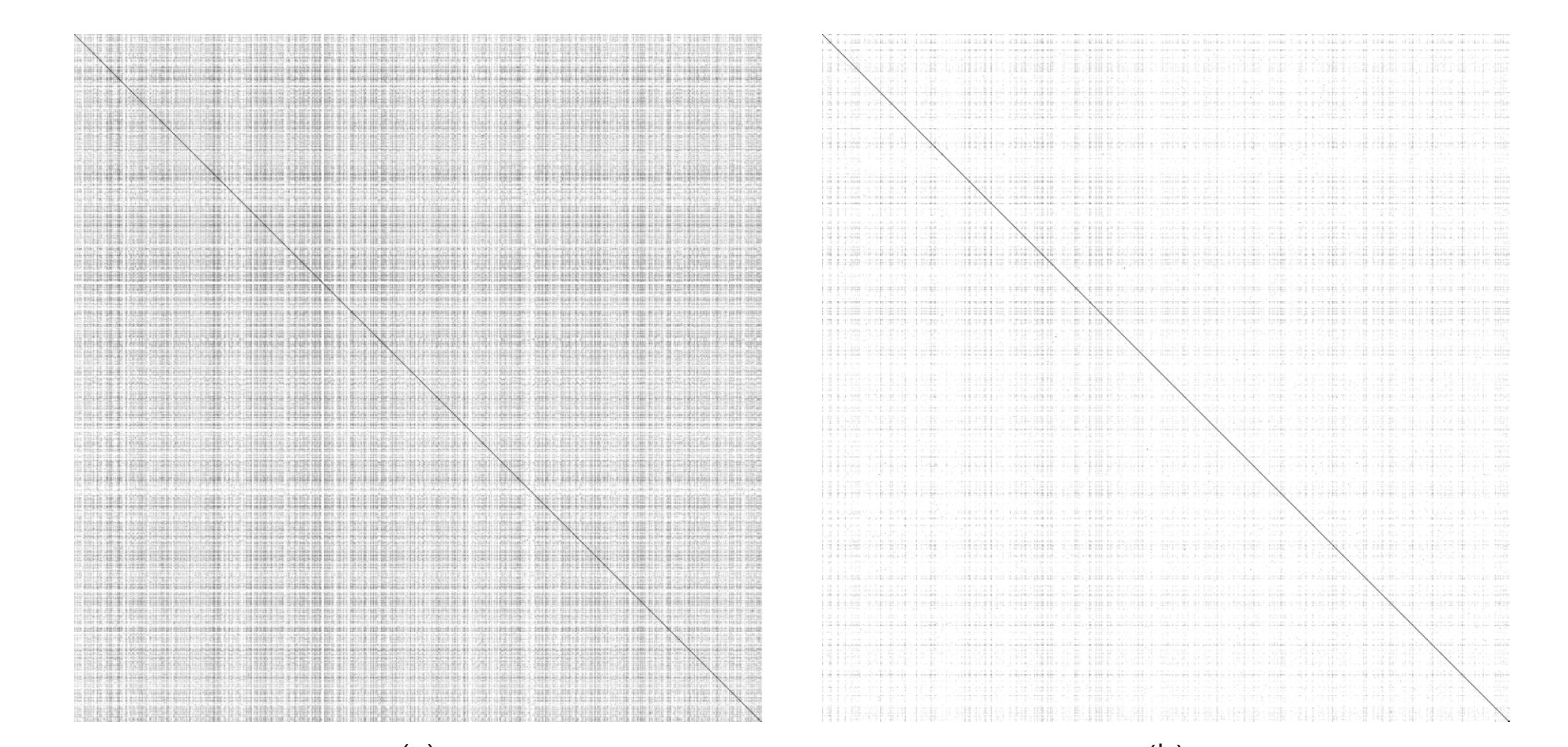}}
  \caption{Visualization of pairwise Euclidean distances among orthogonal projections of unseen classes on ImageNet dataset. These pairwise distances are obtained by using different seen/unseen splits. (a) Results obtained on split 50/950. (b) Results obtained on split 100/900. Darker colors depicts closer distances.}
  \label{fig:imagenet_cm}
\end{figure*}

To demonstrate the necessity of the proposed pre-inspection step, we first split all classes into seen/unseen at different ratios. Then we extract the orthogonal projection of unseen classes on the subspace $\SM$ spanned by seen class embeddings. Finally, we compute the Euclidean pairwise distances among all these projections. These pairwise distances on CUB and ImageNet datasets are visualized in Fig.~\ref{fig:cub_cm} and Fig.~\ref{fig:imagenet_cm}.

We observed that when the number of seen classes is much smaller than that of unseen classes, a lot of pairwise distances tend to 0. This means $f_s$ learned from seen classes is difficulty to discriminate among these unseen classes.

\subsection{Comparison to the state-of-the-art methods}
In addition to Tab. 3 of the main text, we display more details about the experimental setup of these methods in Tab.~\ref{tab:comp-state}.

\begin{table*}[!htp]
\begin{threeparttable}
\centering
\caption{cZSR ($\UM \rightarrow \UM$) comparison on AwA, CUB and Dogs. We compare ours (achieved using 2 iteration) with the state-of-the-art results using different $\KM$, including word vector (W) and attribute (A). We only display the dimension of word vectors in the `Dim of $\KM$' column. In our DMaP, only \emph{skipgram} is used for W. `L' denotes low-level features. `T' or `I' denotes transductive or inductive methods. `+' indicates the concatenation operation.  `--' means no result reported in the original paper. }
  \begin{tabular}{c c c c c c c c c}
  \hline
  Methods & $\XM$ & Dim of $\XM$ & $\KM$ & Dim of $\KM$& T/I &AwA & CUB & Dogs\\
  \hline \hline
  SSE \cite{zhang2015classifying}              & vgg                    & 4096         & A                   &-           &I  &76.23 & 30.41 & --\\
  SJE \cite{akata2015evaluation}               &goog                   & 1024         &  A/W              & 1000   &I  & 66.7 & 50.1 & 33.0 \\
  SynC \cite{changpinyo2016synthesized} & goog          &  1024 & A+W                   & 100          &I  & 72.9 & 54.7 & --\\
  LatEm \cite{xian2016latent}                   &goog                   & 1024          &  A+W +H$^{\ast}$  & 1000   &I  & 76.1  & 51.7& 36.3\\
  RKT  \cite{wang2016relational}              & vgg+goog          & 2024          & A+W             &  500    &I  & 82.43 & 46.24& 28.29 \\
   AMP \cite{fu2015zero}                         & OverFeat            & 4096         & A+W               &  100   &I  & 66     & -- & -- \\
  \hline
  \multirow{2}{*}{TMV-HLP \cite{fu2015transductive} }    & OverFeat & 4096      &A+W                &  1000  &T  & 73.5 & 47.9 & -- \\
                                                                      & OverFeat + DeCaF & 8192             & A+W & 1000         & T  & 80.5& - & - \\
   UDA \cite{kodirov2015unsupervised}    & OverFeat            & 4096          & A                   &  -       &T  &75.6  & 40.6 & --\\
   PST \cite{rohrbach2013transfer}           & L                        & 10940          & A                    & -       & T  & 42.7 & -- & --\\
  \hline\hline

\multirow{15}{*}{\textbf{DMaP}}     & \multirow{3}{*}{OverFeat} & \multirow{3}{*}{4096} & A & - & T & 80.35 & 51.01 & -\\
                                      &  &  & W & 500 & T & 68.80 &26.02 &- \\
                                      & &  & A+W &500 & T & 83.50 & 50.8 &- \\
\cline{2-9}
                                   & \multirow{3}{*}{vgg}        & \multirow{3}{*}{1000}           & A    & -         & T          & 85.66 &  50.45 & --\\
                                       &                      &            &W    & 500     &T         &  82.78 & 23.31 & 33.57 \\
                                        &                    &            &A+W& 500     & T         & \textbf{87.62} & \textbf{52.14 }& -- \\
\cline{2-9}
                                     & \multirow{3}{*}{goog}                     & \multirow{3}{*}{1024}          & A   & -         & T         & 74.94 &  \textbf{61.79} & --\\
                                       &                      &         &W   & 500      &T         & 67.90  & 31.55 & 38.92 \\
                                        &                    &          &A+W& 500     & T        & \textbf{78.61} & 59.62 & -- \\
\cline{2-9}
                                      &\multirow{3}{*}{res}         & \multirow{3}{*}{1000}           & A   & -         & T         & 89.34 & 59.28 & --\\
                                      &                      &         &W   & 500      &T         & 85.70  & 29.97 & 40.18 \\
                                        &                   &          &A+W& 500     & T        & \textbf{90.15} & \textbf{60.90} & -- \\
\cline{2-9}
                                       &\multirow{3}{*}{vgg+goog}            & \multirow{3}{*}{2024}         & A   & -          & T        & 87.52 &\textbf{ 63.79} & --\\
                                       &               &        &W   & 500       &T       & 75.03 & 30.34 & 44.59 \\
                                        &              &         &A+W & 500     & T     & \textbf{91.52} & 62.62 & -- \\
 \hline
  \end{tabular}
  \begin{tablenotes}
  \item[1] OverFeat and DecaF  denote deep features extracted from OverFeat \cite{sermanet2013overfeat} and DeCaF \cite{donahue2014decaf}.
  \item[2] $^{\ast}$ Results obtained by using two types of word vectors, \ie word2vec and glove. H denotes hierarchical semantic embeddings derived from WordNet.
  \end{tablenotes}
  \label{tab:comp-state}
  \end{threeparttable}
\end{table*}

\newpage

\newpage

\end{document}